\documentclass{article}
\usepackage{spconf,amsmath,graphicx,subcaption,multirow,xcolor,colortbl,makecell}
\usepackage{hyperref}
\usepackage{fancyhdr}
\hypersetup{
	colorlinks=true,
	linkcolor=cyan,
	filecolor=blue,      
	urlcolor=red,
	citecolor=green,
}
\usepackage[T1]{fontenc}
\definecolor{lightgreen}{rgb}{0.75, 0.95, 0.75}

\title{JointRF: End-to-End Joint Optimization for Dynamic Neural Radiance Field Representation and Compression}
\name{
Zihan Zheng\textsuperscript{*}, Houqiang Zhong\textsuperscript{*}, Qiang Hu\textsuperscript{\dag},  Xiaoyun Zhang\textsuperscript{\dag}, Li Song, Ya Zhang, Yanfeng Wang
 \thanks{\textsuperscript{*} Authors contributed equally to this work. \textsuperscript{\dag} Corresponding Authors (email:qiang.hu@sjtu.edu.cn, xiaoyun.zhang@sjtu.edu.cn). This work is supported by National Natural Science Foundation of China (62271308), STCSM (22511105700, 22DZ2229005), 111 plan (BP0719010), and State Key Laboratory of UHD Video and Audio Production and Presentation. 
 \\© 20XX IEEE. Personal use of this material is permitted. Permission from IEEE must be obtained for all other uses, in any current or future media, including reprinting/republishing this material for advertising or promotional purposes, creating new collective works, for resale or redistribution to servers or lists, or reuse of any copyrighted component of this work in other works.}
}

\address{Shanghai Jiao Tong University}
\begin{document}
%\ninept
%
\maketitle
\begin{abstract}
% Neural Radiance Fields (NeRFs) excel in photo-realistically  static scenes, inspiring numerous efforts to facilitate volumetric videos. However, rendering dynamic and long-sequence radiance fields remains challenging, due to the significant amount of data required to represent volumetric videos. In this paper, we propose a novel end-to-end joint optimization scheme of dynamic NeRF representation and compression, called JointRF, thus achieving significantly improved compression efficiency against the previous methods. Specifically, JointRF employs a compact residual feature grid along with a coefficient feature grid to represent the dynamic NeRF. This representation handles large motions without compromising quality, while concurrently diminishing temporal redundancy. Additionally, we introduce a sequential feature compression scheme to further reduce spatial-temporal redundancy. Finally, both the representation and compression subnetworks are end-to-end trained in a combined manner within the JointRF.  Extensive experiments demonstrate that JointRF can achieve superior compression performance across various datasets.

Neural Radiance Field (NeRF) excels in photo-realistically static scenes, inspiring numerous efforts to facilitate volumetric videos. However, rendering dynamic and long-sequence radiance fields remains challenging due to the significant data required to represent volumetric videos. In this paper, we propose a novel end-to-end joint optimization scheme of dynamic NeRF representation and compression, called JointRF, thus achieving significantly improved quality and compression efficiency against the previous methods. Specifically, JointRF employs a compact residual feature grid and a coefficient feature grid to represent the dynamic NeRF. This representation handles large motions without compromising quality while concurrently diminishing temporal redundancy. We also introduce a sequential feature compression subnetwork to further reduce spatial-temporal redundancy. Finally, the representation and compression subnetworks are end-to-end trained combined within the JointRF. Extensive experiments demonstrate that JointRF can achieve superior compression performance across various datasets.

\end{abstract}

%both of which are end-to-end trained in a combined manner within the JointRF.
%This feature grid is transformed into coefficients, which are then quantized and entropy encoded,
%通过将长序列动态辐射场在时间上表征为多个特征序列组，每个组包含一个长期参考特征和后续若干该参考帧的残差特征。随后，在特征域对该表征进行了量化和熵编码，并通过端到端的联合优化，使得动态辐射场表征具有低熵性。We represent the scene as a long-term reference feature and the residual features of subsequent frames based on that reference feature.Specifically

%rendering dynamic, long-duration radiance fields remains challenging, due to require a significant amount of data to represent volumetric videos.

% separates the modeling from compression process, resulting in suboptimal compression efficiency.
%
\section{Introduction}\label{sec:intro}

% The photo-realistic volumetric video provides an immersive experience in virtual reality and telepresence. The dynamic Neural Radiance Field (NeRF)\cite{nerf} has demonstrated significant potential in representing photo-realistic volumetric video. However, there are still challenges in storing and transmitting volumetric video using NeRF, especially for arbitrary motions and long-duration sequences. The difficulty lies in identifying an efficient representation and compression method for dynamic NeRF to deliver and store long sequences.
% zhonghq
Photo-realistic volumetric video offers an immersive experience in virtual reality and telepresence. Dynamic Neural Radiance Field (NeRF)\cite{nerf} has shown significant potential in representing photo-realistic volumetric video. However, there are still challenges in storing and transmitting volumetric video using NeRF, especially for sequences involving arbitrary motions and long durations. The difficulty lies in identifying an efficient representation and compression method for dynamic NeRF to deliver and store lengthy sequences.

%Realistic dynamic free-view 3D video has very important research value because of its powerful realism and interactivity, which allows it to bring an immersive experience to the audience. 3D video has huge application potential in the metaverse, virtual reality and various other fields, and has been developing rapidly in recent times. However, current technology still has many limitations. First, the dynamic neural radiation field method can only express very limited segments of dynamic scenes and does not support streaming. This makes it very difficult to efficiently model longer sequences of free-viewpoint dynamic scenes. At the same time, the memory required to model 3D or even 4D scenes increases exponentially, which brings huge difficulties in the storage and transmission of models. Therefore, modeling and compression methods suitable for long sequence dynamic scenes are urgently proposed.

\begin{figure}[t]
\includegraphics[width=\linewidth]{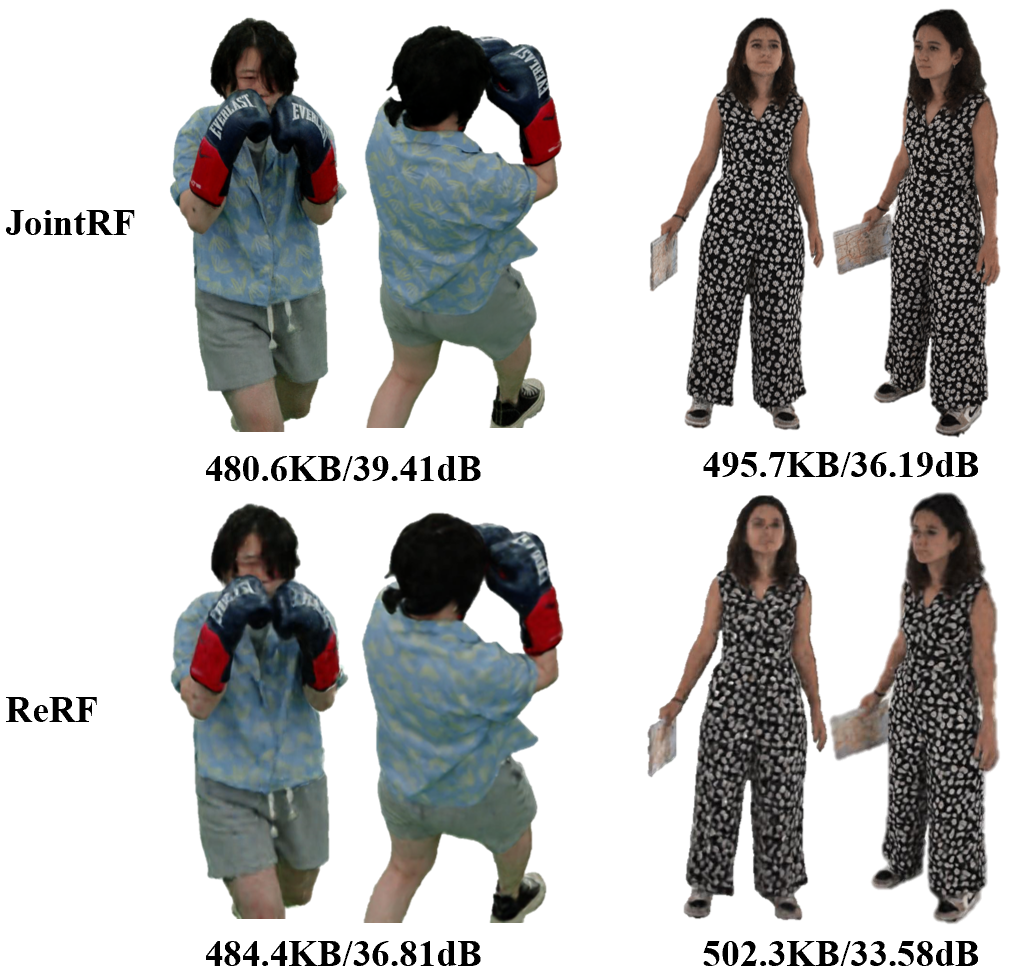}
\caption{The reconstructed results of our proposed JointRF compared to ReRF\cite{rerf} on various datasets.} \label{teaser}
\end{figure}
% zhonghq 修改第二段，nerf效果好；
% 静态nerf扩展到动态nerf，工作简介，运动motion grid 、拓扑问题分析；
% 基于4D动态建模的方法，从压缩角度分析问题，存储空间大 kplanes, 
% 动态压缩工作简介，VQRF
% 动态辐射场压缩rerf分析,自然图像jpeg
% 我们的研究思路 
NeRF and its variants\cite{nerf,instant-ngp,merf} have achieved significant success in synthesizing novel views, surpassing existing 3D reconstruction methods. 
The state-of-the-art visual quality has inspired numerous derivative research studies on dynamic scenes. 
Directly extending per-frame static NeRF methods to dynamic scenes is impractical, as they neglect the spatiotemporal continuity of scenes, resulting in an excess of network parameters.
Some methods\cite{Nerfies,DNeRF,NeuralRadianceFlow} attempt to recreate features in each frame by warping them back into canonical space. However, relying solely on the canonical space restricts the effectiveness of sequences with significant motion or topological changes. 

Other approaches\cite{HumanRF,tineuvox,kplanes} extend the radiance field to 4D spatial-temporal domains, yet these methods face suboptimal rendering quality and encounter challenges related to large model storage, especially in streaming scenarios. Limited efforts have focused on compressing the dynamic radiance field for streaming, with ReRF\cite{rerf} being a representative work. 
Although ReRF\cite{rerf} has achieved some redundancy reduction in dynamic features through the traditional image encoding method, the traditional image encoder is unsuitable for high-dimensional feature domains. Additionally, ReRF\cite{rerf} does not jointly optimize the representation and compression of the radiance field, leading to the loss of dynamic details and a decrease in compression efficiency.

In this paper, we present JointRF, a novel approach that jointly optimizes the representation and compression of dynamic NeRF, achieving better quality and higher compression efficiency (see Fig.\ref{teaser}). Inspired by DiF \cite{factorfields}, the radiance field is decomposed into a coefficient feature grid and a basis feature grid. JointRF explicitly models the residual feature grid between the long-reference basis and the non-long-reference basis.  JointRF only utilizes a long-reference basis and a coefficient grid to represent the first keyframe. In contrast, for each subsequent frame, a compact residual grid along with a coefficient grid is used to compensate for errors and newly observed regions. A major advantage of this representation is the full utilization of feature relevance between adjacent frames.

Moreover, the feature grids of the representation are sequentially quantized and entropy encoded to further reduce redundancy. More importantly, we conduct end-to-end joint optimization for both the representation and compression processes, significantly enhancing the rate-distortion (RD) performance. Considering the non-differentiability of quantization and entropy coding in the compression process, we use simulated quantization and entropy model-based bitrate estimation to facilitate end-to-end training. Experimental results show that JointRF outperforms the state-of-the-art methods in terms of RD performance. To summarize, our contributions include:

\begin{itemize}
    \item We propose a novel end-to-end learning scheme, called JointRF, that can jointly optimize both dynamic NeRF representation and compression. Our approach achieves superior RD performance and eliminates the need for intricate multi-stage training.
    \item We present an efficient and compact representation, representing the 4D radiance field into a series of residual feature grids to support streamable dynamic and long-sequence radiance field.
    \item We introduce an entropy-minimization compression method to guarantee the radiance field features with low entropy. 
\end{itemize}

\section{Related work}
\label{sec:format}

%\subsection{Novel-view Synthesis Methods for Static Scenes}
%Neural radiance fields (NeRF)\cite{nerf} offer photorealistic 3D modeling, yet face long rendering times due to intense computations. Subsequent advancements transitioned scene representation from MLPs to more explicit structures like 3D grids\cite{dvgo,dvgo+}, hash tables\cite{instant-ngp}, and others\cite{mobilenerf,plenoxels,tensorf}, which reduces reconstruction time but increased storage needs. Further developments introduced more efficient representations and reduced feature dimensionality\cite{snerg,merf}, and the advent of factor-fields\cite{factorfields} summarizes many expressions and led to the novel DiF representation. While static scene reconstruction is now efficiently achieved, reconstructing dynamic 3D sequences with growing storage demands continues to be a significant challenge.

\subsection{Dynamic Radiance Field Representation}
Generating realistic synthesized views becomes more challenging in dynamic scenes, particularly due to the presence of moving objects.
%Dynamic scenes necessitate advanced neural representations for efficient, fluid precise motion capture and data compactness. 
% Simple methods to modeling the scene frame by frame using static methods can lead to difficulty in reducing motion blur and excessive redundancy. 
% To maintain spatial integrity and accelerate reconstruction, dynamic NeRF frameworks, like\cite{kplanes,tensor4d,fpo++,tineuvox,streaming} have been innovated.
The canonical space methods\cite{Nerfies,DNeRF,NeuralRadianceFlow} recover temporal features by warping the live-frame space back into the canonical space, which is fragile to large motions and topological changes. Another category of methods\cite{HumanRF,tineuvox,kplanes,tensor4d,fpo++,tineuvox,streaming} extends the radiance field to 4D spatial-temporal domains, where they model the time-varying radiance field in a higher-dimensional feature space.
ReRF\cite{rerf} adopts the residual radiance field technique by leveraging compact motion grids and residual feature grids to exploit inter-frame feature similarity, achieving favorable outcomes in representing long sequences of dynamic scenes.
% However, this approach introduces constraints, such as the necessity for sequential operations like rendering, which may impact performance. 
% Optimizing dynamic 3D reconstruction thus remains a critical and challenging task.
\subsection{NeRF Compression}
%After substantial development, deep learning-based compression methods \cite{compressiclr2017,variational,mentzer2020high} have matured. These methods employ an entropy model to estimate feature distributions for compression, aiming to minimize the compressed bitstream size. 

%Recently, NeRF has integrated compression methods to reduce storage usage, with techniques\cite{vqrf,ecrf} focusing on static NeRF compression. 
%VQRF\cite{vqrf} compresses static radiance field models using entropy encoders, whereas ECRF\cite{ecrf} maps radiance field features to the frequency domain before applying entropy encoding. 
%ECRF\cite{ecrf} incorporates DCT transformation and entropy estimation during training for static scenes. 
%Now, dynamic scene compression remains less explored. As for the compression of the dynamic radiance field, ReRF\cite{rerf} separates compression from training, utilizing a traditional image encoder, which is not suitable for feature grid compression. Overall, there is still a clear demand for end-to-end deep learning compression techniques tailored to dynamic 3D scenes, a gap that our method efficiently addresses.

Recently, there has been limited attention to compressing NeRF to reduce storage usage. VQRF\cite{vqrf} employs an entropy encoder to compress the static radiance field model, while ECRF\cite{ecrf} maps radiance field features to the frequency domain before applying entropy encoding. However, these methods remain confined to static scenes, and there has been relatively little exploration in compressing dynamic scenes. ReRF\cite{rerf} is designed for modeling dynamic scenes and utilizes traditional image compression methods for feature compression. However, due to the separate representation and compression processes, ReRF\cite{rerf} lacks end-to-end optimization. Moreover, traditional image encoders are unsuitable for compressing feature grids, resulting in suboptimal compression efficiency. Comparing to \cite{zhiyu}, our approach introduces different representation and compression methods.

\section{Method}
\label{sec:method}
%In this segment, we delineate the details of JointRF. Initially, for a given 3D scene, our JointRF deploys Basis expansion to represent the scene utilizing two grid types: coefficient and basis. Subsequently, JointRF undertakes direct training of these grids for key frames, while for non-key frames, training is conducted via the residual basis grid technique. Although the basis operates at multiple resolutions, for simplicity, it will be depicted as a grid in this context. In this process, JointRF express a continuously dynamic scene as a group. Post the estimation of quantitative losses, the grid is processed to evaluate the compression rate and extract salient features. Following this, the interpolated feature values are multiplied and channeled through a compact MLP. The culmination of this process is the execution of volume rendering, which yields the final image. 

In this section, we introduce the details about the proposed JointRF representation for long-sequence dynamic scenes (Sec.\ref{sec:representation}), followed by an end-to-end joint optimization scheme for representation and compression (Sec.\ref{sec:optimization}).

%This includes our Dynamic Neural Radiance Field Modeling Based on Residuals (section 1), overall framework (section 2) and Variable bitrate end-to-end compression(section 3).
%符号是什么含义要讲清楚，x=xxxx  d等于什么
%\subsection{Residual-Based Long Dynamic Neural Radiance Field Modeling}
\subsection{JointRF Representation}
\label{sec:representation}
Recall that NeRF\cite{nerf}  employs an implicit function to represent scenes. This function uses a large multi-layer perceptron (MLP) to map spatial coordinate $\mathbf{x} = (x, y, z)$ and view direction $\mathbf{d}$ to color $\mathbf{c}$ and density $\sigma$.  By accumulating the colors $\mathbf{c}_i$ and densities $\sigma_i$ of all sampled points along a ray $\mathbf{r}$, we can derive the predicted color $\hat{\mathbf{c}}(\mathbf{r})$ for the corresponding pixel:
%The pixel color $\hat{C}(\mathbf{r})$ is then estimated by sampling and integrating along the light path.
\begin{equation}
    \hat{\mathbf{c}}(\mathbf{r})=\sum_i^N T_i(1-exp(-\sigma_i\delta_i))\mathbf{c}_i,
\end{equation}
where  $T_i = exp(-\sum^{i-1}_{j=1}\sigma_i\delta_i)$, and $\delta_i$ denotes the distance between adjacent samples.

To maintain high efficiency in training and rendering, we adopt an explicit representation similar to previous work \cite{factorfields}. 
The representation of a static scene is decomposed into the coefficient feature grid $\mathbf{C}$  and the basis feature grid $\mathbf{B}$ via basis expansion. The basis feature grid captures signal commonality, while the coefficient feature grid depicts spatial variations. The radiance field of a static scene can then be expressed as:
%we decompose a continuous signal characterizing a 3D space into two components via basis expansion: coefficients and basis.

%In the above equation, $\delta_i$ represents the distance between consecutive samples, while $T_i = exp(-\sum^{i-1}_{j=1}\sigma_i\delta_i)$. Drawing inspiration from prior work\cite{factorfields}, we decompose a continuous signal characterizing a 3D space into two components via basis expansion: coefficients and basis. %Through the combination of these two parts, the 
%original signal can be restored very closely.
%And the procedure to derive $c$ and $\sigma$ can be expressed as follows:
\begin{equation}
(\mathbf{c},\sigma)= \Phi(interp(\mathbf{x},\mathbf{C}) \circ interp(\mathbf{x}, \mathbf{B} ),\mathbf{d}),
\label{eq2}
\end{equation}
where $\Phi$ is a tiny MLP, and $interp(\cdot)$ represents the interpolation function on the grids.  $\circ$ denotes a Hadamard product.
%where $\mathbf{B}$ is the basis grids of the signal, structured to support multi-resolution configurations, while $\mathbf{C}$ is the coefficient grid of the signal.  $p$ is the number corresponding to the grids and $\Phi$ is a compact MLP. Consequently, a 3D scene at time $t$ can be succinctly represented as $\mathbf{f}_t = \{\mathbf{B}_t,\mathbf{C}_t\}$.

\begin{figure}[t]
\includegraphics[width=\linewidth]{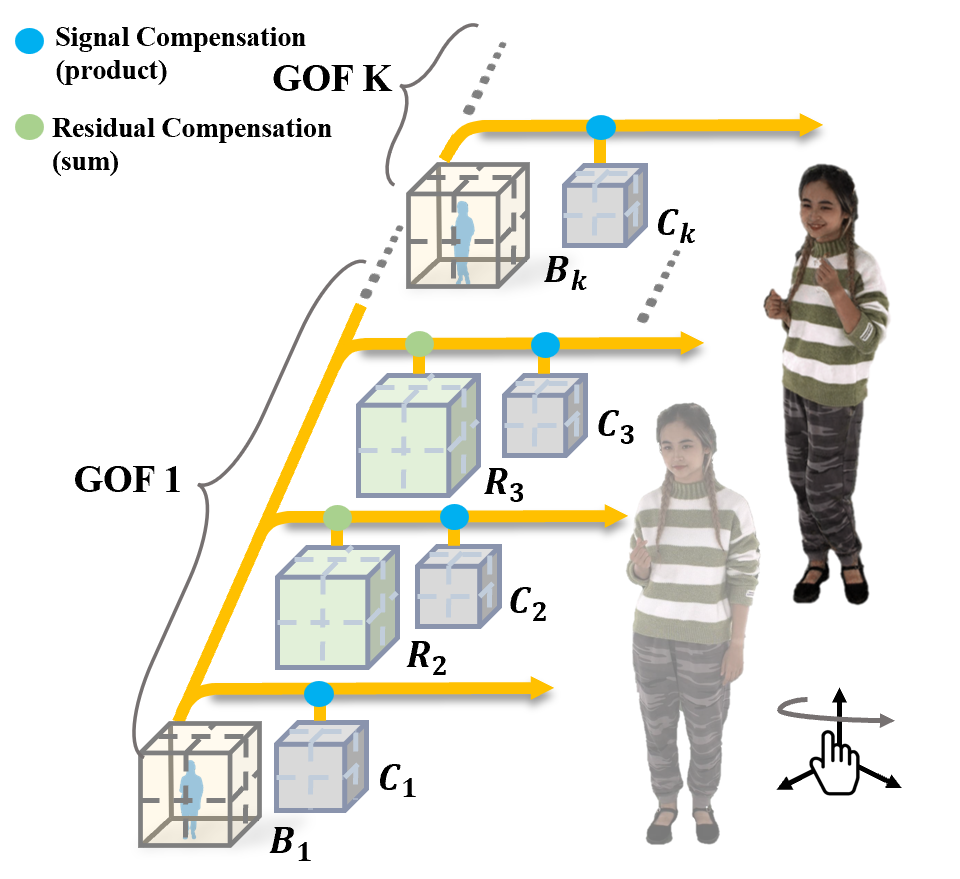}
\caption{Overview of JointRF representation. We divide the sequence into several GOFs. Each GOF starts with a keyframe, represented by a long-reference basis feature grid and a coefficient feature grid. Each subsequent frame in the GOF is then represented by a residual feature grid $\mathbf{R}_t$ in conjunction with a coefficient feature grid $\mathbf{C}_t$.} \label{group}
\end{figure}

When extending the radiance field from a static scene to a dynamic one, a simplistic approach might involve employing individual per-frame feature grids $\{\mathbf{B}_t, \mathbf{C}_t\}_{t=1}^N$ for the dynamic scene consisting of $N$ frames. However, this method fails to account for crucial temporal coherence. This not only leads to discontinuities in visual quality during significant motion but also results in excessive data volume due to the presence of substantial temporal redundancy, posing a particularly severe issue for long sequences in dynamic scenes.

To enhance temporal continuity while minimizing redundancy, we propose a highly compact representation method. The overview of our representation method is illustrated in Fig.\ref{group}. We represent the long sequence of dynamic radiance field over time as multiple continuous groups of feature grids (GOF), which is a collection of successive grids. Each GOF $\mathbf{G}=\{\mathbf{f}_t\}_{t=1}^N$  begins with a keyframe $ \mathbf{f}_1 = \{\mathbf{B}_1,\mathbf{C}_1\}$, which is a long reference feature grid. Due to the commonality captured by the basis in signal extraction and the short-term similarity of features, each remaining frame in the GOF is represented as a compact residual grid along with a coefficient grid ($\mathbf{f}_t=\{\mathbf{R}_t,\mathbf{C}_t\}$, $1<t\leq N$). The residual feature grid $\mathbf{R}_t$ indicates the compensation for errors and newly observed regions between the keyframe's basis grid and the current frame's basis grid. $\mathbf{R}_t$ helps enhance the representation by incorporating residual information and reducing temporal redundancy. Once obtained $\mathbf{R}_t$, we can calculate the basis feature grid of the $t$-th frame by adding the residual compensation:$\mathbf{B}_t = \mathbf{B}_1 + \mathbf{R}_t$, enabling the recovery of the current radiance field by applying the signal compensation according to Eq.(\ref{eq2}).
It is noteworthy that $\mathbf{C}_t$ is the coefficient based on $\mathbf{B}_t$, and learning the residuals of $\mathbf{C}_t$ holds no physical significance.
% Once $\mathbf{R}_t$ is obtained, the basis feature grid of the $t$-th frame can be recovered by adding the residual compensation: $\mathbf{B}_t = \mathbf{B}_1 + \mathbf{R}_t$. This process enables the reconstruction of the current radiance field, applying the signal compensation according to Eq.(\ref{eq2}).
%The residual feature grid $R_t$ indicates the residual compensation for errors and newly observed regions between the basis grid in keyframe and  the basis grid  the current frame.

%We represent the long sequence of dynamic radiance fields over time as multiple feature sequence groups, with each group containing a long-term reference feature and subsequent residual features for that reference frame.

%Note that our JointRF representation enables highly efficient sequential feature modeling. Given the keyframe's basis feature grid $\mathbf{B}_1$, the basis feature grid of the $t$-th frame can be obtained from $\mathbf{R}_t$:

Note that our JointRF representation enables efficient sequential feature modeling with several key advantages. Firstly, it leverages the simplicity of the residual data distribution, making it easier to compress and transmit compared to the complete data distribution. Secondly, our representation handles large motions without compromising quality. Additionally, since our method relies on only one keyframe during rendering, it enables simultaneous rendering of multiple non-keyframes, facilitating parallel computation. Furthermore, this representation supports frame-by-frame loading and rendering, reducing memory usage and favoring streaming.

\begin{figure*}
\includegraphics[width=\linewidth]{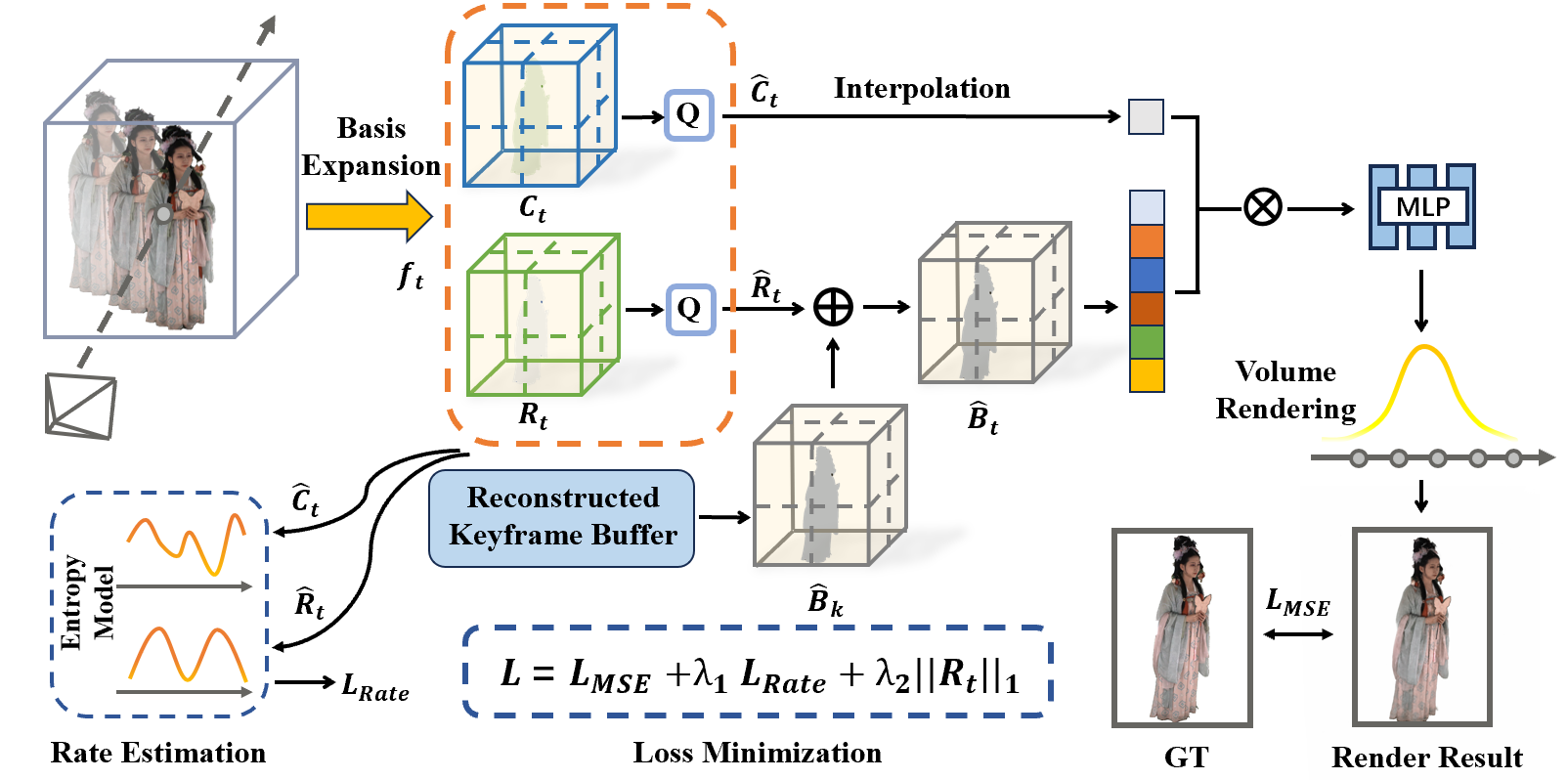}
\caption{Overview of our JointRF training. First, we apply simulated quantization to generate $\mathbf{\hat C}_t$ and $\mathbf{\hat R}_t$, and then estimate the rate of $\mathbf{\hat C}_t$ and $\mathbf{\hat R}_t$ as loss during the forward pass. Next, we load the long-reference basis feature grid $\mathbf{\hat B}_k$ from the reconstructed keyframe buffer and combine it with $\mathbf{\hat C}_t$ and $\mathbf{\hat R}_t$ to obtain the MSE loss. Finally, we sequentially train each frame and update the residual feature grid, coefficient feature grid, and the corresponding entropy model.}

%of the training process of  JointRF. First, we represent dynamic 3D scenes with coefficient and residual basis grids and then quantize. Next, we feed the quantized grids into an entropy model for rate estimation to get $L_{rate}$ and combines $\hat R_{t}$ and $\hat B_k$ from key frame buffer for $\hat B_t$. Finally, the grid undergoes interpolation, is multiplied, and rendered to produce the result and get $L_{MSE}$.} 
\label{train}
\end{figure*}
\subsection{End-to-end Joint Optimization}
\label{sec:optimization}
%Here, we introduce a compr

Here, we introduce an end-to-end optimization approach that jointly optimizes the representation and compression of dynamic radiance field to further improve compression efficiency. The overall framework of our proposed approach is illustrated in Fig.\ref{train}. We apply simulated quantization to the feature grid $\mathbf{f}_t$ and utilize an entropy model-based bitrate estimation to facilitate end-to-end training. The objective of JointRF is to ensure that the radiance field representation learned from modeling has low entropy while maintaining high reconstruction quality.

%To minimize the memory demands of 3D scene representation, we employ an end-to-end joint optimization approach, merging the training of neural radiance fields with the compression model. This strategy diverges from conventional techniques, which typically optimize the two parts separately and then integrate them. Instead, we address this challenge holistically, optimizing each pipeline component in unison to achieve both superior compression rates and enhanced reconstruction quality.

\textbf{Simulated Quantization.} The quantization operation can effectively reduce bitrate of the feature grids during the compression process, but it also results in a certain degree of information loss. 
Introducing quantization operation during training can effectively enhance the model's robustness to the quantization loss. 
However, the rounding operation disrupts gradient propagation and makes it incompatible with end-to-end training.
Since quantized value $\mathbf{Q}(x) \in [x-\frac{1}{2},x+\frac{1}{2}]$,  we introduce uniform noise within the range of $[-\frac{1}{2},\frac{1}{2}]$ to simulate the quantization effect, as shown in Eq.(\ref{equ:quantization}), enabling gradient propagation. 
% Consequently, this technique ensures that the average quantization loss of grids is minimized.

\begin{equation}
\mathbf{Q}(x) = x+u, u \sim U(-\frac{1}{2},\frac{1}{2}). \label{equ:quantization}
\end{equation}

\textbf{Rate Estimation.} 
We perform entropy encoding on the quantized feature grids to generate a highly compressed bitstream. 
If acquiring the bitrate during the training phase is possible, this metric could be integrated into the loss function to encourage a lower-entropy distribution of features, effectively imposing a bitrate constraint on the network updates.
However, entropy encoding does not preserve gradients like the quantization process. To address this, we introduce an entropy model within the training phase to estimate the entropy of the grids, which is the lower bound of the bitrate after compression. The entropy model can approximate the probability mass function (PMF) of quantized $\hat y$ of a feature grid by computing the cumulative distribution function (CDF) of $\hat{y}$, as shown in Eq.(\ref{PMF}). 
\begin{equation}
P(\hat{y}) = P_{CDF}(\hat{y}+\frac{1}{2}) - P_{CDF}(\hat{y}-\frac{1}{2}). \label{PMF}
\end{equation}

To maintain precision, we refrain from presuming any preconceived data distribution for the 3D grids. Instead, we construct a novel distribution within the entropy model to closely match the actual data distribution. %Throughout the training process, both coefficient and basis grids are inputted into the entropy model. 
During training, the entropy model predicts the size of the compressed bitstream as part of the overall loss. The training process of JointRF can be seen in Fig.\ref{train}. 

On the encoding side of compression, we employ quantization followed by entropy encoding, specifically a rangecoder to compress the features and get the bitstream. On the decoding side, entropy decoding is performed to recover the features. The entire procedure unfolds as follows:
\begin{figure*}[]
\includegraphics[width=\linewidth]{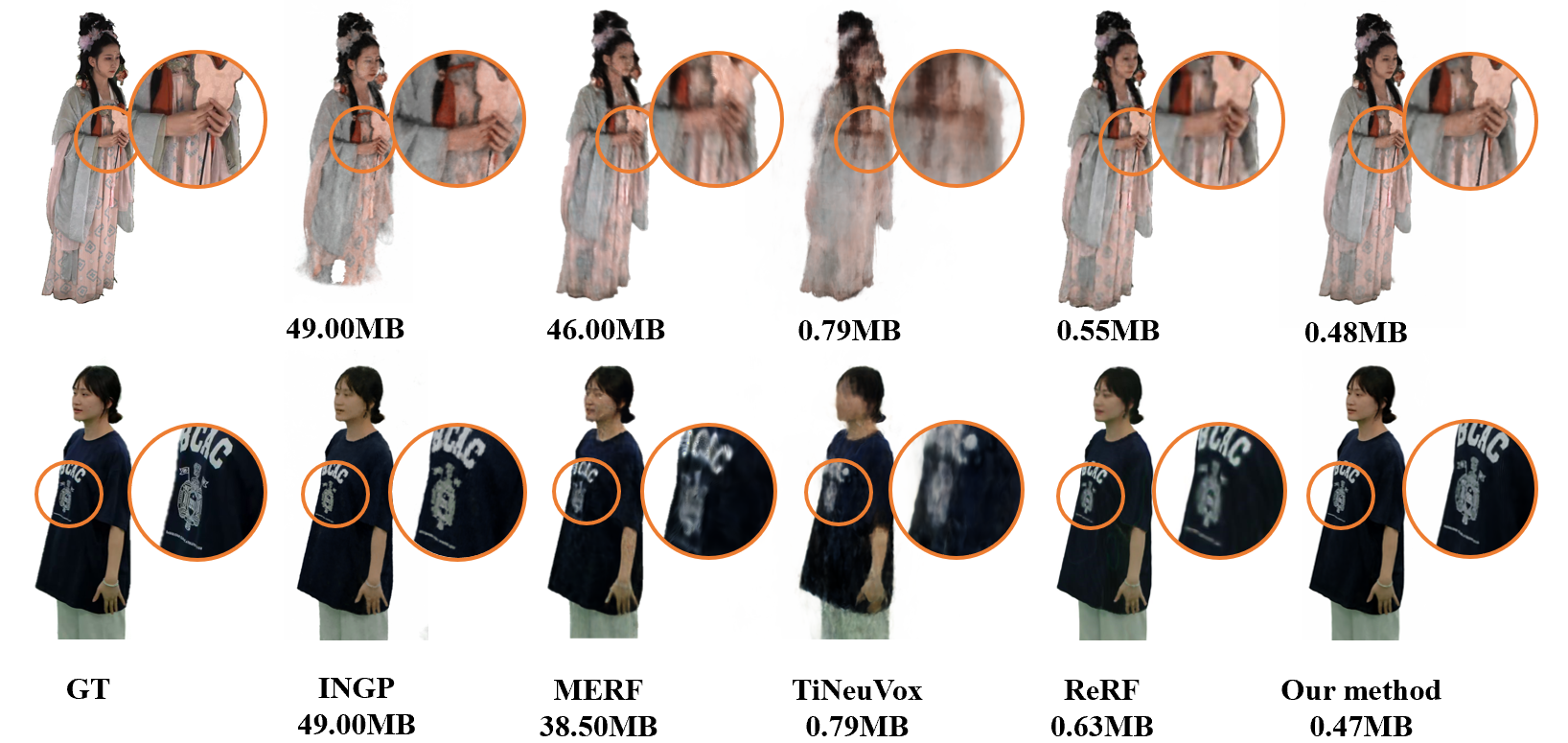}
\caption{Qualitative comparison against dynamic scene reconstruction methods and per-frame static reconstruction methods.} \label{qualitative}
\end{figure*}
%By leveraging end-to-end joint optimization, we acquire features characterized by low entropy and an enhanced reconstruction efficacy. These attributes ensure that the features not only facilitate precise scene reconstruction, but are also readily compressible. 
\begin{equation}
\hat{x}=\frac{\mathbf{D}(\mathbf{E}\left(\mathbf{Q}(q\cdot x)-\mathbf{Q}\left(q\cdot \min(x)\right)\right))+\mathbf{Q}(q \cdot \min(x))}{q},
\end{equation}
where $\mathbf{E}$ is the entropy encoder, $\mathbf{D}$ is the entropy decoder and $\mathbf{Q}$ represents the quantization operation. To enable compression into int8 format, we convert the compressed data to non-negative values and subsequently revert it to its original range during decompression. The variable $q$ is the quantization parameter. During quantization, data is multiplied by $q$, effectively expanding the data range, which, in turn, subtly enhances quantization precision. By adjusting the parameter $q$, a trade-off can be made between reconstruction quality and model storage. A higher value of $q$ yields improved model reconstruction quality, but also increases the size of the compressed model.

\textbf{Training Objective.}
The total loss function of the entire system can be written as below:
\begin{align}
    \mathcal{L}_{total} &= \mathcal{L}_{mse} + \lambda_1 \mathcal{L}_{rate} + \lambda_2 ||\mathbf{R}_t||_1 ,\\
\mathcal{L}_{mse} &= \sum||\mathbf{c}(\mathbf{r})-\hat{\mathbf{c}}(\mathbf{r})||^2 ,\\
\mathcal{L} _{rate} &= -\frac{1}{N} \sum_{\hat{y} \in \hat{ \mathbf{R}}_t }{\log_2\left(P_1(\hat{y})\right)} - \frac{1}{M} \sum_{\hat{y} \in \hat{ \mathbf{C}}_t }{\log_2\left(P_2(\hat{y})\right)},
\end{align}
where $\mathcal{L}_{mse}$  
metrics the difference
between the ground truth and the result rendered by JointRF.
$\mathcal{L} _{rate}$ represents the estimated rate derived from $\hat{\mathbf{R}}_t$ and $\hat{\mathbf{C}}_t$. $||\mathbf{R}_t||_1$ is L1 regularization applied to $\mathbf{R}_t$ to ensure temporal continuity and minimize the magnitude of $\mathbf{R}_t$ . The parameter $\lambda_1$ is used to balance the rate and distortion, allowing for control over the model size and reconstruction quality. The parameter $\lambda_2$ measures the extent of our constraint on $\mathbf{R}_t$.
 
\section{Experimental Results}
\subsection{Configurations}
\textbf{Datasets.} In this section, we extensively assess JointRF through experiments conducted on five sequences: two from the ReRF\cite{rerf} dataset and three from the DNA-Rendering\cite{dna} dataset. The ReRF dataset comprises 74 camera views, of which we designate 70 for training and the remaining 4 for testing. Images in this dataset have a resolution of $1920 \times 1080$. The DNA-Rendering dataset, on the other hand, includes 48 views, with 46 used for training and 2 for testing. Images in this dataset have a resolution of $2048 \times 2448$. 

\noindent \textbf{Setups.} During the quantization phase, we typically evaluate four distinct $q$ values:  1, 2, 5, 10. For loss function, we initialize both $\lambda_1$ and $\lambda_2$ at 0.000001 and allow $\lambda_1$ decrementing progressively. For feature fitting, we employ seven distinct entropy models, each corresponding to the dimensions of the six different-resolution basis grids and the single coefficient grid. In our experimental setup, the length of each GOF is customarily fixed at 10.

%\subsection{Experimental Configurations}

\subsection{Comparison}
To the best of our knowledge, JointRF is a unique approach for jointly training dynamic NeRF and their associated compression processes in a unified end-to-end manner. To validate the effectiveness of our approach, we compare with several state-of-the-art methods for dynamic scenes including INGP\cite{instant-ngp}, MERF\cite{merf}, TiNeuVox\cite{tineuvox}, ReRF\cite{rerf} both qualitatively and quantitatively.  In Fig.\ref{qualitative}, we present the visual results of two sequences. As illustrated, our method demonstrates discernible superiority in both the compactness of the model size and the precision of detail rendering not only on per-frame static reconstruction method INGP\cite{instant-ngp} and MERF\cite{merf} but also on dynamic scene reconstruction methods TiNeuVox\cite{tineuvox} and ReRF\cite{rerf}.

\renewcommand{\arraystretch}{1.2}
\setlength{\fboxsep}{1pt}
\setlength{\tabcolsep}{2pt}
\begin{table}[]
\scalebox{0.84}{ 
\begin{tabular}{c|c|cc|cc|c}
%\multicolumn{1}{l}{}                           & \multicolumn{1}{l}{}                          & \multicolumn{1}{l}{} & \multicolumn{1}{l}{\colorbox{lightgreen}{best}} & \multicolumn{1}{l}{} & \multicolumn{1}{l}{}       & \multicolumn{1}{l}{}   \\ 
\hline

\multirow{2}{*}{Dataset} & \multirow{2}{*}{Method} & \multicolumn{2}{c|}{Training View} & \multicolumn{2}{c|}{Testing View} & \multirow{2}{*}{\makecell{Size\\(MB)}$\downarrow$}  \\ \cline{3-6}
                         &                         & PSNR $\uparrow$          & SSIM $\uparrow$           & PSNR $\uparrow$          & SSIM $\uparrow$         &                       \\ \hline
\multirow{5}{*}{ReRF}                    & MERF\cite{merf}                    & 32.99          & 0.981          & 27.33          & 0.964         & 38.5                \\
                         & INGP\cite{instant-ngp}                   & 36.48          & 0.989          & 28.57          & 0.966         & 49.0                \\
                         & TiNeuVox\cite{tineuvox}                & 30.19          & 0.962          & 22.39          & 0.950         & 0.81               \\
                         & ReRF\cite{rerf}                    & 37.03          & 0.990          & 30.04          & 0.977         & 0.58               \\ \cline{2-7} 
                         & Ours                    & \textbf{38.68}          & \textbf{0.992}          & \textbf{33.23}          & \textbf{0.980}         & \textbf{0.47}               \\ \hline
\multirow{5}{*}{\makecell{DNA-\\Rendering}}     & MERF\cite{merf}                    & 33.24          & 0.975          & 27.52          & 0.958         & 46.0                \\
                         & INGP\cite{instant-ngp}                   & 34.13          & 0.979          & 24.22          & 0.960         & 49.0                \\
                         & TiNeuVox\cite{tineuvox}                & 30.28          & 0.953          & 22.15          & 0.945         & 0.80              \\
                         & ReRF\cite{rerf}                    & 34.28          & 0.981          & 29.39          & 0.977         & 0.60               \\ \cline{2-7} 
                         & Ours                    & \textbf{35.17}          & \textbf{0.982}          & \textbf{31.83}          & \textbf{0.978}         & \textbf{0.48}               \\ \hline
\end{tabular}
}
\caption{Quantitative comparison against dynamic scene reconstruction methods and per-frame static reconstruction methods. We calculate the average PSNR, SSIM, and storage for each frame across all training and testing views, separately.}
\label{t1}

\end{table}
\setlength{\tabcolsep}{5pt}
%As for quantitative comparison, we adopt three metrics to evaluate our method, including peak signal-to-noise ratio (\textbf{PSNR}), structural similarity index (\textbf{SSIM}), and the average model storage for each frame. As depicted in Table \ref{t1}, 

We also conduct a quantitative comparison in terms of Peak Signal-to-Noise Ratio (\textbf{PSNR}), Structural Similarity Index (\textbf{SSIM}), and model storage as shown in Table \ref{t1}. It can be seen that our method outperforms other methods and achieves the best reconstruction quality with the lowest model storage. INGP\cite{instant-ngp} requires a large amount of storage load when modeling 3D scenes, and cannot model unknown perspectives well. Although MERF\cite{merf} can reduce the storage load through baking operations, it still requires dozens of MB, and the reconstruction effect is not satisfactory. TiNeuVox\cite{tineuvox}, on the other hand, is capable of representing sequences of arbitrary length within a consistent memory footprint. However, it suffers from severe blurring effects as the frame count increases. Compared to ReRF\cite{rerf}, our method performs better in terms of PSNR, SSIM, and model storage as well. 

%Our method, in comparison, surpasses ReRF\cite{rerf} in terms of reconstruction quality while maintaining a similar level of storage usage. 

%Table \ref{t2} shows the comparisons of RD performance between our JointRF and ReRF\cite{rerf}.  In the Compared to ReRF, our method demonstrates substantial advantages across the two datasets, particularly on the ReRF dataset.

Table \ref{t2} shows a comparison of RD performance between our JointRF and ReRF\cite{rerf}. Notably, our JointRF achieves a better RD performance compared to ReRF. On the ReRF dataset, we observe average BDBR reductions of 51.28\% and 69.66\% for training and testing views, respectively. Similarly, on the DNA-Rendering dataset, the average BDBR saving is 35.06\% and 36.38\% for training and testing views, respectively. The superior performance of our method can be attributed to the fact that ReRF\cite{rerf} employs a traditional encoder, which is unsuitable for feature grid compression. In contrast, JointRF conducts an end-to-end joint optimization of representation and compression, leading to enhanced RD performance.

\begin{table}[]
\scalebox{0.84}{
\begin{tabular}{c|cc|cc}
\hline
\multirow{2}{*}{Dataset} & \multicolumn{2}{c|}{Training View} & \multicolumn{2}{c}{Testing View} \\ \cline{2-5} 
                         & \makecell{BD-PSNR\\(dB)} $\uparrow$     & \makecell{BDBR\\(\%)} $\downarrow$       & \makecell{BD-PSNR\\(dB)}  $\uparrow$   & \makecell{BDBR\\(\%)}  $\downarrow$     \\ \hline
ReRF                     & 3.12            & -51.28     & 5.18           & -69.66    \\ \hline
DNA-Rendering            & 1.88            & -35.06     & 1.96           & -36.38    \\ \hline
\end{tabular}
}
\caption{ The BDBR results of our JointRF when compared with ReRF\cite{rerf} on different datasets.}
\label{t2}
\end{table}
% zhonghq train view下的平均PSNR差异只有1.88， 和其他几幅图差距较大，需要解释原因

\subsection{Ablation Studies}
% FacterFields baseline(no compress, no residual) 一个点
% Full: Facterfields + encoder + residual module + End2End 红色曲线
% 消融残差模块: Facterfields + encoder - residual module ，蓝色曲线
% 消融编码器：facterfields + residual - encoder 一个 点
% 消融End2End: facterfields + residual + encoder 绿色曲线
% rerf 黄色曲线
We conduct three ablation studies on both training and testing views on ReRF and DNA-Rendering datasets to validate the effectiveness of each component in our method by removing each of them individually. We mainly focus on the compression module, end-to-end joint optimization, and dynamic residual representation.
% to validate the effectiveness of each component in our method.
%We show the effi of  compression module by removing the entropy module.
%As shown in Fig1 point, 

%specifically focusing on the compression module, end-to-end joint optimization and dynamic residual representation by removing each of them individually from the full method. 
%Additionally, this representation does not inherently integrate any form of compression.

In the first ablation study, we removed the compression step while preserving the residual-based dynamic modeling. In the second experiment, we did not jointly optimize dynamic modeling and compression during training and performed compression after training completion. Finally, we individually modeled dynamic scenes frame by frame and conducted joint optimization without introducing residual-based representation.

\begin{figure}[t]
  \centering
  \begin{subfigure}[b]{0.49\linewidth}
    \includegraphics[width=\linewidth]{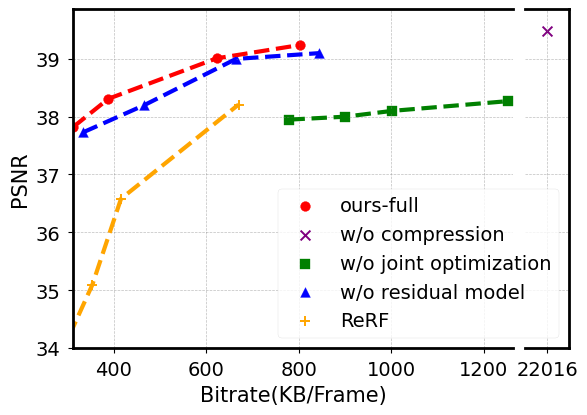}
    \caption{ReRF (Train) }
    \label{fig:sub1}
  \end{subfigure}
  \hfill
  \begin{subfigure}[b]{0.49\linewidth}
    \includegraphics[width=\linewidth]{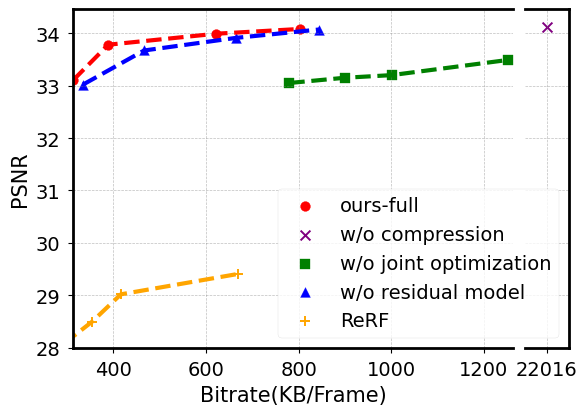}
    \caption{ReRF (Test)}
    \label{fig:sub2}
  \end{subfigure}
  
  \begin{subfigure}[b]{0.49\linewidth}
    \includegraphics[width=\linewidth]{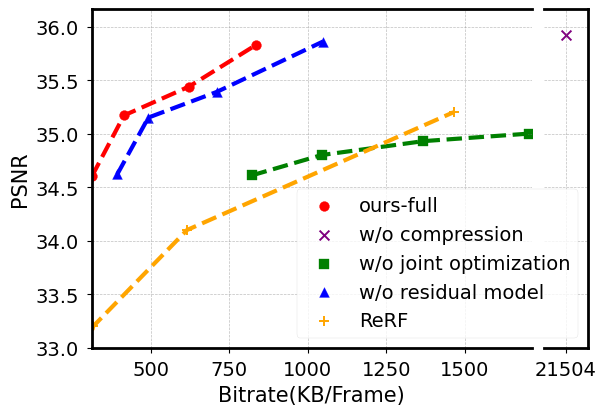}
    \caption{DNA-Rendering (Train)}
    \label{fig:sub3}
  \end{subfigure}
  \hfill
  \begin{subfigure}[b]{0.49\linewidth}
    \includegraphics[width=\linewidth]{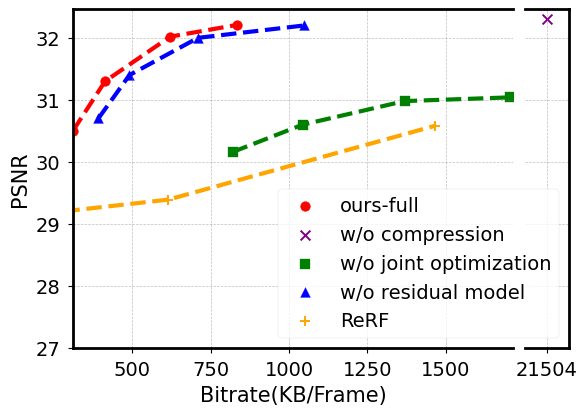}
    \caption{DNA-Rendering (Test)}
    \label{fig:sub4}
  \end{subfigure}
  
  \caption{ Rate-distortion curves in both the ReRF and DNA-Rendering datasets. Rate-distortion curves not only illustrate the efficiency of various components within our JointRF but also demonstrate its superiority over ReRF\cite{rerf}. }
  %Ablation studies on individual components and performed quantitative Rate-Distortion comparisons with ReRF on the ReRF and DNA-Rendering datasets.}
  \label{rd}
\end{figure}

The results of the ablation studies can be seen in Fig.\ref{rd}. It shows that our compression method significantly reduces model storage by approximately 40 times  while maintaining comparable reconstruction quality. What's more, joint optimization can not only  reduce the model size but also slightly improve PSNR because the features obtained through joint optimization are easier to compress and robust to quantization errors. Lastly, dynamic residual representation can effectively achieve better RD performance. The results of the ablation studies demonstrate the integral importance of our dynamic residual representation, compression module, and joint optimization strategy.

%Subsequently, we assessed the advantages of our end-to-end joint optimization. By comparing our model with a counterpart lacking this optimization, we demonstrated that our method can improve the quality of compressed model reconstruction while enabling higher compression ratios. This is because joint optimization can make our features have low entropy characteristics and be robust to quantization errors. Lastly, we compare our method with the approach of independently training each static frame without dynamic residual modeling. The findings from this comparison underscored that dynamic residual representation can effectively achieve better RD performance.

\section{Conclusion}
\label{sec:conclusion}

In this paper, we propose JointRF, a novel approach that jointly optimizes the representation and compression of dynamic NeRF. We first introduce a highly compact modeling method for representing dynamic and long-sequence NeRF.
To further reduce the spatial-temporal redundancy, we devise a compression method that is concurrently optimized with the representation of dynamic NeRF, enabling end-to-end training. Instead of predetermining feature distributions, our approach models data distributions during training to enable precise bitrate estimations and quantitative differentiable approximations. Experimental results show that JointRF outperforms the state-of-the-art methods in terms of RD performance across various datasets. 
With its unique representation and compression capabilities for long-sequence dynamic scenes, we believe our approach lays the foundation for various potential applications in volumetric videos.

%In this study, we introduce a neural radiance field modeling technique apt for long-sequence dynamic scenes. We have devised a compression method that is concurrently optimized with the representation of dynamic NeRF, facilitating end-to-end training. Instead of predetermining feature distributions, our approach models data distributions during training to enable precise bitrate estimations and quantitative differentiable approximations. Moreover, we segment the scene into multiple groups and employ the concept of residuals to learn non-key frames based on key frames, enhancing learning efficiency and reconstruction quality.

%The primary limitations of our work include the lack of further compression of the MLP and the absence of dynamic group size adjustment. Additionally, the inability to dynamically alter the compression ratio during rendering might pose challenges in practical applications. These areas present promising avenues for future research and development in the field.

%\vfill
%\pagebreak

\bibliographystyle{IEEEbib}
\bibliography{refs}

\end{document}